\documentclass[letterpaper, 10 pt, conference]{ieeeconf}

\IEEEoverridecommandlockouts                              

\usepackage{graphicx}
\usepackage{caption}
\usepackage{booktabs}
\usepackage{subcaption}
\usepackage{array}
\usepackage{float}
\usepackage{makecell}
\usepackage{mathtools, amsmath}
\usepackage{algorithm}
\usepackage{algpseudocode}
\usepackage{relsize}
\usepackage{lscape}
\usepackage{multicol, multirow}
\usepackage{tablefootnote}

\usepackage[utf8]{inputenc}
\usepackage{parskip}
\usepackage{listings}
\usepackage{amsmath}
\usepackage{gensymb}

\title{\LARGE \bf
Towards Increasing the Robustness of Predictive Steering-Control Autonomous Navigation Systems Against Dash Cam Image Angle Perturbations Due to Pothole Encounters*
}

\author{Shivam Aarya$^{1}$%
\thanks{*This work was not supported by any organization}%
\thanks{$^{1}$Department of Computer Science, Johns Hopkins University, Baltimore, MD 21218, USA 
        {\tt\small saarya1@jhu.edu}}%
}

\begin{document}

\maketitle
\thispagestyle{empty}
\pagestyle{empty}

\begin{abstract}

Vehicle manufacturers are racing to create autonomous navigation and steering control algorithms for their vehicles. These software are made to handle various real-life scenarios such as obstacle avoidance and lane maneuvering. There is some ongoing research to incorporate pothole avoidance into these autonomous systems. However, there is very little research on the effect of hitting a pothole on the autonomous navigation software that uses cameras to make driving decisions. Perturbations in the camera angle when hitting a pothole can cause errors in the predicted steering angle. In this paper, we present a new model to compensate for such angle perturbations and reduce any errors in steering control prediction algorithms. We evaluate our model on perturbations of publicly available datasets and show our model can reduce the errors in the estimated steering angle from perturbed images to 2.3\%, making autonomous steering control robust against the dash cam image angle perturbations induced when one wheel of a car goes over a pothole.

\end{abstract}

\section{Introduction}

Cars have become the main transportation method in many parts of the world. An average American drives 14,263 miles per year, amounting to 3.2 trillion miles driven annually, according to the Federal Highway Administration~\cite{meyer_2023}. Increasing adoption of cars around the world also drives the increase in the number of road accidents. As per the Centers for Disease Control and Prevention (CDC), 1.35 million people are killed annually due to road accidents, making it the eighth leading cause of death. While analysis of road safety focuses on driving under the influence, speeding, mobile phone use, and fatigued driving~\cite{pires2020car}, they usually overlook the danger posed by deteriorating road conditions leading to the formation of potholes. 

Cars are increasingly being equipped with automated steering control, autopilot (i.e., autonomous braking and acceleration), and predictive safety maneuvers, towards full self-driving capabilities. By reducing the dependence on human drivers, they can reduce the leading causes of accidents that are all attributed to driver errors. However, these autonomous navigation systems have not yet been adapted to the problem of avoiding potholes on the road that human drivers usually handle. The automated systems will simply drive at full speed without adjusting the steering angle (the degree to which the steering wheel of the car is rotated) over a pothole as if it does not exist, which can be extremely dangerous if an inattentive driver fails to take control and correct the car's trajectory before the impact.

While most drivers consider them a minor nuisance that makes their drive less comfortable or a minor obstacle on the road to be avoided, potholes can actually have a dangerously significant impact on a vehicle. Direct contact with a pothole could result in an impact that causes injury, major damage to the vehicle, or even a result equivalent to that of a 35 mph vehicular crash~\cite{bodden_2021}. This is not an uncommon problem, with \$26 billion being spent by drivers in 2021 to pay for pothole repair, and with 1 in 10 drivers needing to repair their vehicle after hitting a pothole~\cite{edmonds_2022}. 

Therefore, it is becoming crucial to address the issue of pothole avoidance in these systems and to predict a car's response to driving over a pothole. While research is being done to develop methods to recognize~\cite{kaushik2022pothole} and avoid potholes~\cite{raja2022spas}, these capabilities are yet to appear in production vehicles. When an autonomous vehicle goes over a pothole, the images captured by the cameras get perturbed. These perturbations can impact the accuracy of steering control at the moment of a pothole encounter. There have been works to make autonomous steering control models robust to irrelevant objects (e.g., buildings, shrubs, etc.) in the scene not encountered in the training data so as to make pre-trained models robust to unseen irrelevant objects~\cite{wang2020end, shen2021gradient}. However, there have been very few works to develop methods to make autonomous steering control robust to perturbations in images due to the vehicle's encounters with a pothole. It is important to consider the consequences of a pothole impact and to regulate its effect on the car's autonomous navigation systems to maintain a safe response to the impact and keep the driver as far as possible from harm.

In this paper, we train a denoising autoencoder to increase the robustness of the autonomous navigation systems against the image perturbations (any deviation made to the image to change it from the original image, e.g., rotation in dash cam image angle) that result from encounters with potholes. We train and test our proposed model via experiments on publicly available data sets collected in real life.

\section{RELATED WORK}

Multiple researchers have attempted to solve the problem of corrective obstacle avoidance. In these methods, the avoidance logic can be easily extended to potholes in the road such that the autonomous navigation model will turn the steering wheel accordingly to avoid the incoming pothole on the road. The researchers have proposed different approaches that can broadly be categorized into one of three categories: vibration-based, depth-based, and GPS-based approaches.

\subsection{Vibration-Based Approach}

Some works use sensors such as speedometers and accelerometers to identify and classify potholes with vehicle encounters~\cite{mednis2011real, lekshmipathy2021effect, kulkarni2014pothole, eriksson2008pothole}. They do this by mathematically modeling the jerk induced by traveling over a pothole on the car’s suspension and then analyzing the sensor input by feeding it into a Bayes decision classifier to record the proper response from hitting the pothole~\cite {sharma2019pothole, carlos2018evaluation}. Some approaches have even employed various deep learning techniques to analyze the accelerometer sensor data~\cite{varona2020deep, kumar2020modern, pandey2022convolution}. These approaches are fairly accurate in identifying when the vehicle encounters a pothole, but due to the limited reaction time of a vehicle, there is very little the algorithm can do to actually avoid the pothole other than know when it is encountered in real-time~\cite{wang2015real, varona2020deep}. There is also very little the algorithm can do in terms of reacting to the information being provided at the moment due to limited processing time and sensor limitations. The reason for this is that the approach can only formulate a reaction to the pothole once it detects it with the vibrations in the car, at which point it is too late to react properly. While this may be a useful aspect of the pothole collision problem by accurately recording the resulting discomfort levels of hitting a pothole, it cannot solve the initial pothole avoidance problem in its entirety~\cite{li2015road, kim2014review, rishiwal2016automatic, salau2019survey}.

\subsection{Depth-Based Approach}

 To help flush out the limitations of the vibration-based pothole avoidance approaches, other researchers have proposed a depth-based pothole mapping system that can use recorded data from LiDAR sensors and the approximate distance of the surface to identify pavement anomalies \cite{kim2014review, 8300645, 8809907, 8890001}. A minor variation of this method employs the use of stereo vision to capture 3-D road surface data \cite{zhang2014efficient, li2018road, ramaiah2021stereo}. This approach is one of the most reliable at identifying information about the upcoming potholes ahead of time, but a LiDAR sensor is limited to only gathering information about the distance to the pavement in front of the car. For the car to perform reactive and corrective steering, it needs to be aware of its surroundings and make decisions based on inputs from its environment as well. This is the same issue seen in the vibration-based approach proposed by~\cite{kang2017pothole, jog2012pothole}.

\subsection{GPS-Based Approach}

Finally, there have been numerous methods proposed to crowd-source the recording of potholes similar to collecting data for Google Maps. These approaches use sensors in the car to identify when it has passed over a pothole, then it records this data along with GPS coordinates and sends it to a central server \cite{kulkarni2014pothole, rishiwal2016automatic, borgalli2020smart}. This information can then be used to warn other cars that are fitted with the same system of upcoming potholes \cite {desai2019design}. However, this means that the car can only react if another car has already gone over the pothole at full speed and recorded it, and if the pothole is patched or worsened, the car will not have updated information as it will simply avoid the area altogether. This may also lead to some privacy concerns with user GPS data being monitored and transmitted through a third-party server, and this is overall not a feasible or viable real-time reactive approach or final solution~\cite{kamalesh2021intelligent, sharma2019pothole, joubert2011pothole}.

\subsection{The Gap}

Each of these three approaches utilizes a different strategy to predict the correct reaction to a pothole encounter. ~\cite{li2015road} analyzes the impact of the pothole on the car mid-collision, while~\cite{kang2017pothole} uses computer vision to classify and identify the pothole prior to it encountering the car. ~\cite{kamalesh2021intelligent} and~\cite{kulkarni2014pothole} propose GPS-based approaches that can identify the pothole well in advance but cannot give information to the self-driving system until one car goes over the pothole and does not have real-time information about the road situation around it. While there is much development being made on these different strategies to handle the moments prior to the encounter, there is very little research about the post-impact effects on the autonomous navigation system. 

Therefore, the purpose of this research is to develop a technique for increasing the robustness of steering control algorithms in response to image perturbations due to the impact of a car with a pothole.

\section{METHOD}

We take the first steps towards our research goal by developing models on real-life datasets and evaluating our model in simulation. Our work can pave the way to perform real-world experiments where cars with cameras can be used to test various algorithms as the car is driven over potholes.

\subsection{Generating 3D Representation of Potholes}

The first step in analyzing the effects of a pothole collision is to establish what the potholes themselves look like. There is a wide variety of potholes on the roads and the range of possible potholes to encounter is very large. In order to represent the average pothole collision, we must first take a sample of various potholes on an average roadway. To do this, we utilize a publicly available dataset that incorporates novel algorithms for road disparity (or inverse depth) transformation and deploys it on a semantic segmentation network~\cite{8300645, 8809907, 8890001}. This means that the researchers developed a technique to use multiple stereo cameras positioned at varying angles on a moving car to capture 3D visualizations of damage in roadways along with RGB overlays (color images) of the same damaged areas.

\begin{figure}[!htb]
    \minipage{0.14\textwidth}
        \includegraphics[width=\linewidth]{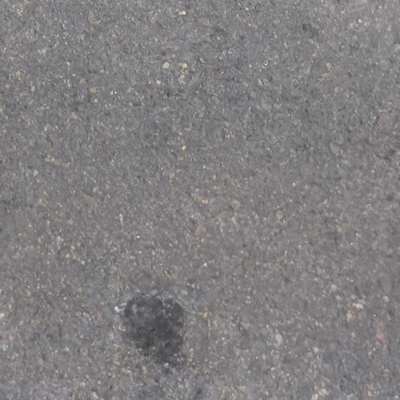}
        \caption{RGB scan of top-down view for a sample pothole}\label{fig:potholeimagergb}
    \endminipage\hfill
    \minipage{0.14\textwidth}
        \includegraphics[width=\linewidth]{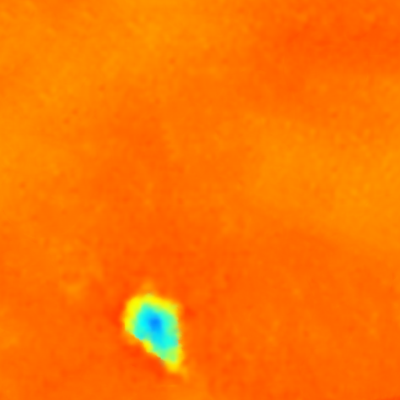}
        \caption{Jet colorscale heatmap of depth for a sample pothole}\label{fig:potholeimagetdisp}
    \endminipage\hfill
    \minipage{0.14\textwidth}%
        \includegraphics[width=\linewidth]{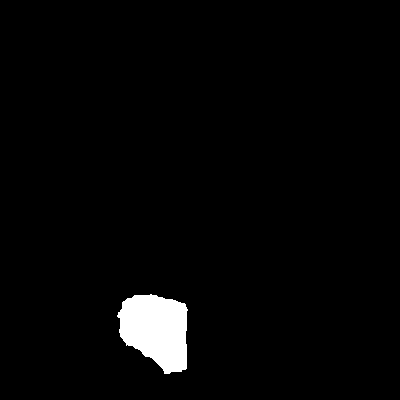}
        \caption{Strictly black-and-white label image for a sample pothole}\label{fig:potholeimagelabel}
    \endminipage
\end{figure}

This dataset includes 600 samples of potholes split into 180 samples in a testing group, 240 samples in a training group, and 180 samples in a validation group. This ratio of images split into multiple groups is commonly used in machine learning applications, but since this research only uses this dataset to represent a sample of all potholes in roadways, we proceed by combining all samples into one group of 600 potholes. This group contains an RGB color image (e.g. Figure~\ref{fig:potholeimagergb}), a heatmap in the jet color scale (e.g., Figure~\ref{fig:potholeimagetdisp}), and a strictly black and white label image (e.g. Figure~\ref{fig:potholeimagelabel}). The heatmap uses the jet color scale because it is the default output color scale to show a displacement map of the pothole (indicating areas of low depth with the color blue and areas of high depths with the color red) in the road for each of the 600 images~\cite{8300645, 8809907, 8890001}.

Next, the pothole images need to be reconstructed into 3D representations of potholes that can then be statistically analyzed to extract various properties of the potholes. These measurements can then be used with measurements from accurate car diagrams to determine the effect of driving over the pothole on a car and its camera.

For this 3-D reconstruction, we use Blender, which is a free and open-source powerful 3D modeling software used by multiple different industries for different parts of the 3D pipeline, such as modeling, sculpting, visual effects, computer graphics, UV Unwrapping, virtual reality, and even creating fully animated films.

Since there are 600 different potholes, it would be a lengthy process to individually combine the three images into a 3D model for each of the samples. Coincidentally, Blender implements a new Python API in Blender 2.5~\cite{blender_2022} that allows for an interaction between the Python programming language and the Blender software. 
We proceeded by creating a Blender script in Python (Algorithm~\ref{alg:blender}) to take each of the 1,800 images (600 groups of 3 images) and recursively process each of them into a complete 3D file. 

\begin{algorithm}
\caption{Blender pothole generation}\label{alg:blender}
    \hspace*{\algorithmicindent} \textbf{input:} RGB, heatmap, and label images for each pothole in the pothole-600 dataset \\
    \hspace*{\algorithmicindent} \textbf{output:} A dataset consisting of a 3-D model for every pothole in the pothole-600 dataset
\begin{algorithmic}

\For{each RGB, heatmap, label image in dataset}
    \State{Add plane mesh}
    \State{Subdivide mesh by 7 divisions}
    \State{Add heatmap as displacement modifier to plane}
    \State{Create new material shader and add RGB image as texture}
    \State{Apply material shader to model}
    \State{Shade smooth the mesh and export as fbx (for 3D point-space data}
    \State{Delete all generated assets to clear workspace for processing next pothole}
\EndFor
\end{algorithmic}
\end{algorithm}

\begin{figure}[h!]
    \centering
    \includegraphics[width=0.45\textwidth]{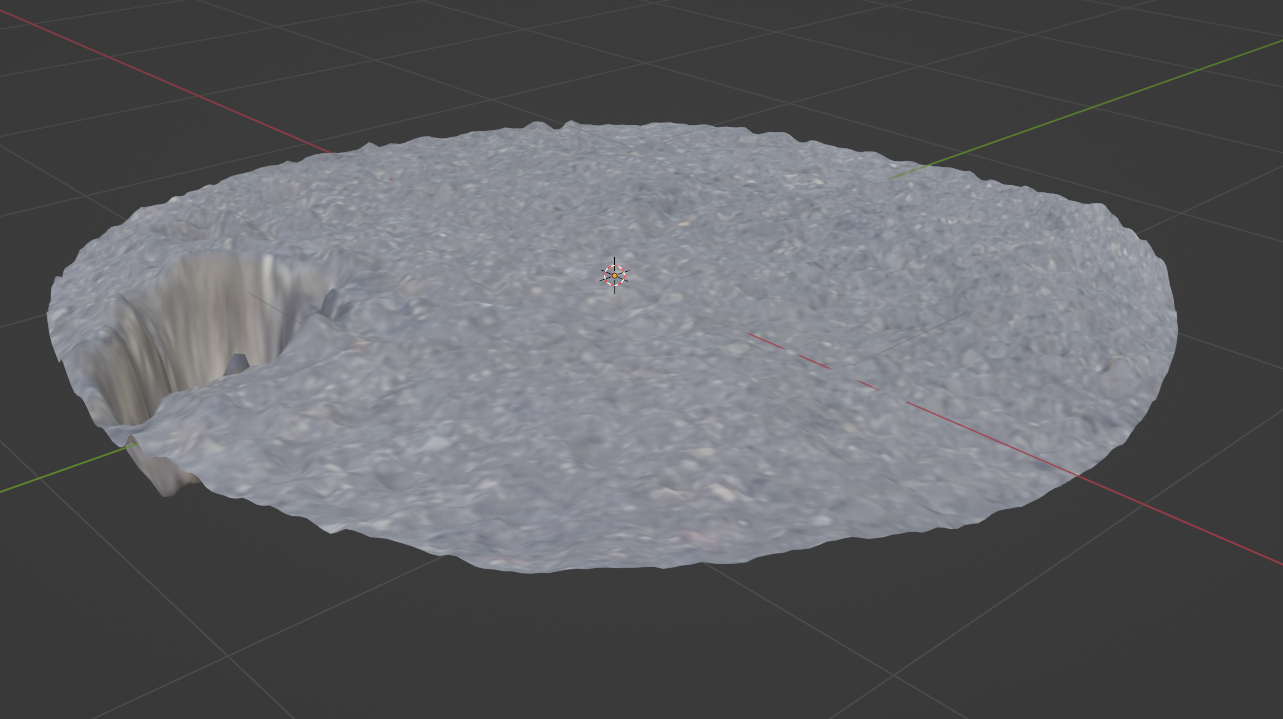}
    \caption{3D Object created by the program in Algorithm~\ref{alg:blender} for a sample pothole}
    \label{fig:pothole3D}
\end{figure}

\subsection{Statistical Characterization of Potholes}

Now that we have a collection of 600 physical models of potholes, the next step is to obtain a statistical representation of those potholes. For this purpose, we developed another Python program to process all of the 3D objects created by the Blender program. This Python program imports each 3D project and then saves a matrix of gradients for every point on the model, essentially creating a dataset of slopes (rate of change in depth of the pothole) for every pixel of the model. Then, the program discretizes this matrix into a $10X10$ set of equally-sized squares, giving us 100 different small portions of the pothole. We then take the average gradient for each of the 100 squares (taking the direction of each of the slopes into account) and save this as the new matrix representing the model of the pothole. The reason we do this is to reduce the amount of data that is needed to represent the pothole, which makes it computationally efficient.

After the program repeats the pothole modeling process for each of the 600 models, it saves the $10X10$ matrix of average gradients of each pothole into one big dataset containing 60,000 values representing each of the potholes. Doing so turns the entire set of 3D pothole models into a low-level representative dataset of numerical values, on which we can then perform a statistical analysis to obtain average statistics and measures of variability and central tendency in the sample of potholes.

\begin{algorithm}
\caption{Statistical 3D pothole analysis}\label{alg:stat}

    \hspace*{\algorithmicindent} \textbf{input:} The dataset of 3-D pothole models generated in Algorithm~\ref{alg:blender} \\
    \hspace*{\algorithmicindent} \textbf{output:} Statistical depth-analysis of all input models

\begin{algorithmic}

\For{each 3D model file}
    \State{Extract 3D point-cloud mesh for depth data}
    \State{Discretize mesh into 100 (10x10) chunks of 40x40 pixels each}
    \State{Calculate average depth gradient for each chunk}
    \State{Calculate average of all chunk gradients from the pothole file}
\EndFor
\State{Perform statistical analysis on all average gradient values found for each pothole}
\end{algorithmic}
\end{algorithm}

\begin{table}[h!]
\centering
\begin{tabular}{||c | c||} 
 \hline
 Statistic & Value (mm)\\  
 \hline\hline
 mean & 61.724417\\
 \hline
 std & 10.208508\\
 \hline
 min & 47.450000\\
 \hline
 25\% & 55.812500\\ 
 \hline
 50\% & 59.490000\\ 
 \hline
 75\% & 63.862500\\ 
 \hline
 max & 108.500000\\ 
 \hline
 \end{tabular}
 \caption{Statistical output from Algorithm \ref{alg:stat} on 600 images}
 \label{table:potholestats}
\end{table}

\subsection{Estimating Camera Angle Perturbation}
Now that we have a statistical representation of potholes, our next step is to estimate the change in the camera angle when the car hits the pothole. For this purpose, we analyze the physical construction of a typical vehicle and treat it as a rigid body, then calculate the effect of hitting the average pothole on the image input from a dash cam mounted facing the front of the car that can be used for autonomous navigation. In order to do this, some basic assumptions about the dimensions of the average car must be made.

In 2022, the best-selling passenger car worldwide was the Toyota Corolla, acquiring 1.12 million sales~\cite{carstats}. Therefore, we use the Toyota Corolla to represent our average, everyday car driving on the roadways. 

\begin{figure}[h!]
    \centering
    \includegraphics[width=0.45\textwidth]{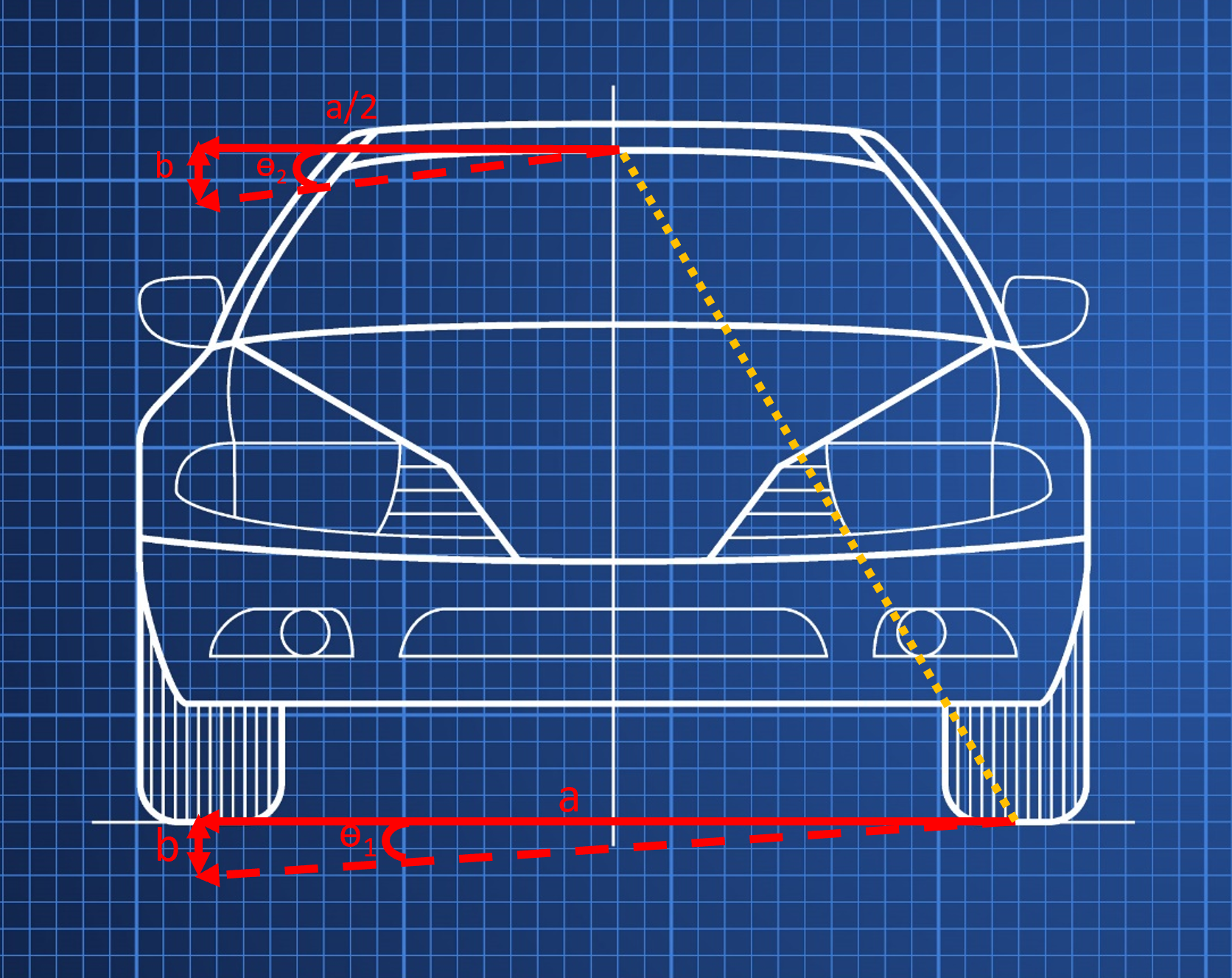}
    \caption{Diagram representing a typical pothole collision with a car \cite{cardiagram}
    }
    \label{fig:cardiagram}
\end{figure}

In Figure~\ref{fig:cardiagram}, $a$ represents the width of the car from the middle of each wheel. This is estimated using the baseline assumption of the Toyota Corolla, which has a width of 1,780 mm (1.78 m). Let $b$ represent the amount that the wheel dips down due to a pothole impact. For the purposes of this research, we assume that the car is a rigid body. Therefore, we can simply use the sample distribution of pothole depths given by our statistical analysis of the pothole models and directly plug them into $b$ to get the offset of the car. After that information is plugged in, we need only to perform some trigonometry to obtain our distribution of angles $\theta_2$ that represents the average dash cam perturbation in terms of the angle by which the image is shifted.

\begin{align*}
\label{eqn:potholeangles}
    \theta_1&= \arctan\frac{b}{a},\theta_2= \arctan\frac{2b}{a} \\
    \theta_2&= \arctan\frac{123.44}{1780}, \text{if } a = 1780 mm, b \approx 61.72 mm \\
    \theta_2&\approx 0.0692 \times \frac{180}{\pi} = 3.967\degree
\tag{Derivation 1}
\end{align*}
\ref{eqn:potholeangles} outlines the steps for converting pothole depths to camera angle perturbations. Next, we perform the steps in~\ref{eqn:potholeangles} on the distribution of pothole angle depths outlined in Table~\ref{table:potholestats} to obtain a distribution of angle perturbations to dashcam images. 

\subsection{Predicting the Steering Angle Correction}
Finally, we develop a steering-angle prediction model that can provide the change in steering angle needed to become resilient against angle perturbations of the camera when the car goes over a pothole.

To represent the images for training a prediction model, we adopt the AutoJoin AutoEncoder~\cite{villarreal2022autojoin}. The basic concept of an autoencoder is to reduce input images to a low-dimensional representation (essentially a blurry image with much less information and detail) and then train the machine learning model to recreate the original image from that low-dimensional representation. Since the model can now recreate an image from one containing less data, this technique is commonly used in denoisers to add detail to an image, but in the case of AutoJoin, it is implemented to increase the robustness of an existing autonomous navigation system against noise in dashcam footage.

AutoJoin trains a denoising autoencoder (DAE) that can predict the steering angle from images that are perturbed due to changes in color, saturation, and brightness. They use a joint loss function that includes errors in the predicted steering angle and the reconstructed image from the perturbed images. We instead focus on angular perturbations in images induced by pothole encounters and hence our loss function only considers the errors in the predicted steering angle.

For this modeling, we use a dataset of dashcam images provided by Audi~\cite{geyer2020a2d2} and Honda~\cite{ramanishka2018toward}, which has about 850,000 training, testing, and validation images. We then apply the distribution of angle perturbations to the training images in this dataset to generate the perturbed training images. For this research, we use the NVIDIA autonomous navigation model (a pre-trained neural network created by NVIDIA to take an input dashcam image and return a predicted steering angle) for steering angle prediction. 

For our autoencoder, the pre-trained model is fed the original dashcam image as the input and the predicted steering angle is regarded as the ground truth or actual correct steering angle in normal conditions. Then, a perturbed dashcam image (as if the car is currently going over a pothole) is fed into the model, and the difference in the predicted output steering angle for the two images (the ground truth steering angle and the perturbed image steering angle) is calculated. 

The autoencoder is trained to reduce the noise in the perturbed image and introduce bias in the model such that the difference between the actual steering angle and the steering angle predicted by the pre-trained model on the perturbed dashcam image is minimized until the model gets as close as possible to the original output even on a perturbed image. Once this happens, the self-driving algorithm has essentially become robust against the image perturbation we have introduced, and will not be nearly as severely affected by the change in dashcam angle introduced when a car is driven over a pothole.

\section{EXPERIMENTS} 
\label{sec:experiments}
We trained and tested our autoencoder on about 500,000 images for 50 epochs (meaning that it goes over each image 50 times when relearning the correct model output) when the model's performance began to stabilize. For our experiments, we used a desktop computer with 32 gigabytes of RAM and an NVIDIA GeForce RTX 3090, which has Tensor cores optimized for processing the large matrices associated with this model training process.

In order to evaluate the performance of the model on the perturbations, we make use of the Mean Squared Error (MSE) in the predicted steering angle estimated from the perturbed image by our model and that determined from the original unperturbed image. 

\begin{figure}[h]
    \centering
    \includegraphics[width=0.45\textwidth]{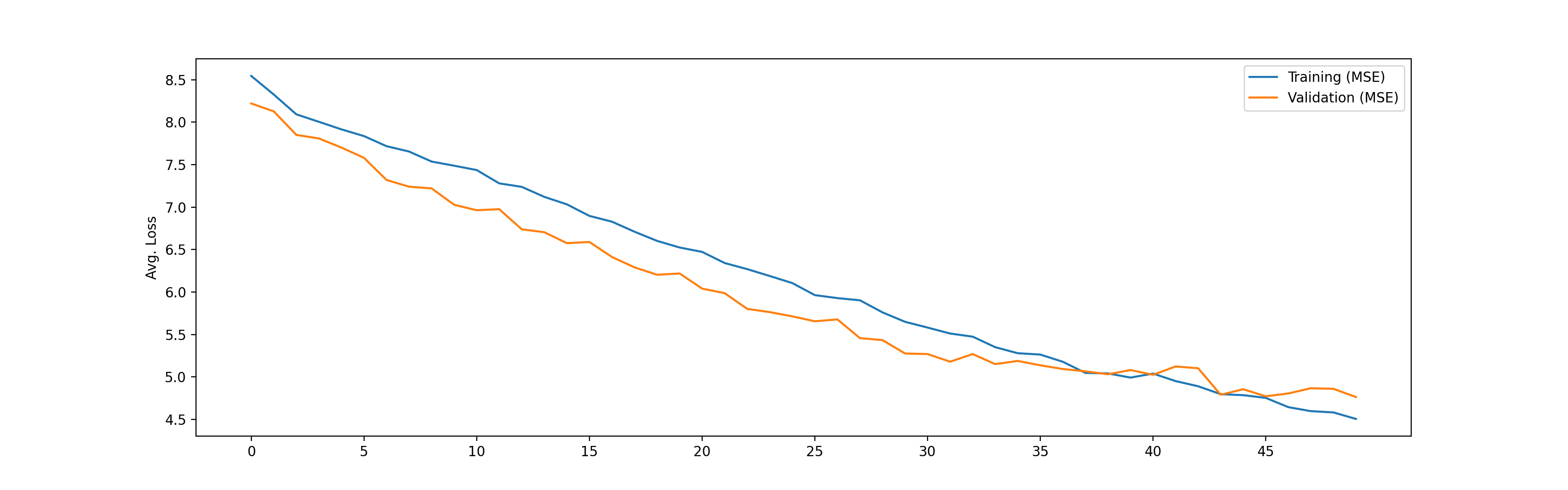}
    \caption{Reduction in the Mean Squared Error (MSE) in the estimated steering angle with increasing epochs}
    \label{fig:traingraph}
\end{figure}

We analyze the change in the MSE after each epoch. As shown in Figure~\ref{fig:traingraph}, our model is able to reduce the validation error in the estimated angle to stabilize at approximately 2.3\% after 50 epochs.

\section{Future Works}

As the experiments show (in Section~\ref{sec:experiments}), our proposed model can be used to make autonomous steering control more resistant to pothole-induced image perturbations. There are many limitations to this work that open up new opportunities to future research.

There were multiple assumptions made in this research that may limit the potential implications of our findings. The most major assumptions were made during the creation of the image perturbation distribution, such as when the car was assumed to be a Toyota Corolla and the car was treated as a rigid body with no suspension. This likely skewed our data to have harsher angles than there would be in real cars, but this was not enough to pose a risk of our model becoming overly resilient. In the real world, the car could be one of a very wide selection of self-driving vehicles and it will likely have a suspension of varying levels of effectiveness. This test is also based almost entirely on simulation, so the technique needs to be implemented in a physical car to test the true impact of the robust model in real-world scenarios. If this were to happen, the car wheelbase and the impact of suspension should be taken into account and incorporated into the experiment to obtain the most accurate results.

The initial pothole sample itself is also a limitation in that it is not a completely representative sample of all potholes, it is only the potholes that the ZED stereo camera was able to catch, so there may be other deformations in the road such as broken speedbumps or major dips that have a similar image perturbation effect on the dash cam footage but were not incorporated in the training of our robust model. While this may skew our results to be less or more severe depending on the frequency of alternative road deformations on the road being tested, since the overall effect on the camera is similar, the robust model should still be able to handle the perturbation despite not being trained and optimized to deal with the scenario.

We only presented the Mean Squared Error (MSE) observed in the proposed model. Future work can compare it with non-trivial baseline methods as well as other alternative approaches and present Average Mean Accuracy Improvement (AMAI)~\cite{shen2021gradient}. 

Ultimately, any model developed for improving autonomous steering control needs to be implemented in existing autonomous navigation systems and tested in real life. Fortunately, the model proposed here can be incorporated into existing autonomous vehicles, without the need for any major restructuring of the preexisting self-driving model.

\section{Conclusion}
This research proposed a new model for creating robustness in an autonomous navigation algorithm to be resilient to perturbations in camera angles at the time of pothole encounters. Improving it further and incorporating it in production vehicles can improve the accuracy of steering control. Even after pothole avoidance becomes a standard feature, it may still not be possible to avoid many potholes due to multiple colocated potholes or not enough margin to maneuver around the pothole in heavy traffic. Therefore, compensating for angle perturbation when encountering potholes will continue to be useful in autonomous vehicles in the foreseeable future.

\bibliographystyle{IEEEtran}
\bibliography{IEEEabrv, IEEEexample}

\end{document}